\documentclass[conference]{IEEEtran}

\usepackage{xspace}
\usepackage{multirow}
\usepackage[overload]{textcase}

\ifCLASSOPTIONcompsoc
 \usepackage[caption=false,font=normalsize,labelfont=sf,textfont=sf]{subfig}
\else
 \usepackage[caption=false,font=footnotesize]{subfig}
\fi

\usepackage{flushend}
\usepackage[table]{xcolor}
\usepackage{amssymb,amsmath,mathtools,fixmath,amsthm}
\usepackage{nicefrac}

\usepackage{cases}

\usepackage{url}

\newcommand{\eg}{e.g.,\xspace}

\renewcommand{\paragraph}[1]{\vspace{2mm}\noindent\textbf{#1}}

\begin{document}
	
\title{Gondola: a Parametric Robot Infrastructure\\for Repeatable Mobile Experiments}

\author{
{Marco Cattani, Ioannis Protonotarios}
\vspace{1.5mm}\\
\fontsize{10}{10}\selectfont\itshape m.cattani@tudelft.nl, i.protonotarios@tudelft.nl\\
\fontsize{10}{10}\selectfont\itshape Delft University of Technology, The Netherlands
}

\maketitle

\begin{abstract}
When deploying a testbed infrastructure for Wireless Sensor Networks (WSNs), one of the most challenging feature is to provide repeatable mobility. Wheeled robots, usually employed for such tasks, strive to adapt to the wide range of environments where WSNs are deployed, from chaotic office spaces to potato fields in the farmland. For this reson, these robot systems often require expensive customization steps that, for example, adapt their localization and navigation system. 

To avoid these issues, in this paper we present the design of Gondola, a parametric robot infrastructure based on pulling wires, rather than wheels, that avoids the most common problems of wheeled robot and easily adapts to many WSN's scenarios. 
Different from wheeled robots, wich movements are constrained on a 2-dimensional plane, Gondola can easily move in 3-dimensional spaces with no need of a complex localization system and an accuracy that is comparable with off-the-shelf wheeled robots.
\end{abstract}

\section{Introduction}
\label{sec:intro}
Providing a repeatable movement is essential for a wide range of WSN's applications, from automated testing to optimal sensor placements. 
To this end, several WSN mobile infrastructures couple wireless sensors with wheeled small-scale robots that are cheap and easily available on the market. Unfortunately, wheeled robots present several drawbacks that practically limit the range of mobile experiments a researcher can run.
Firstly, in order to navigate, affordable wheeled robots often require a localization infrastructure that accurately estimate the robot's position and heading: from simple black lines on the ground ~\cite{Rensfelt2010} to complex tracking systems based on a camera~\cite{Johnson2006}.
Secondly, these mobile robots rely on batteries as source of power, limiting the maximum duration of an experiment and imposing a periodic recharging task.
Finally, and most importantly, wheeled robots can only move on a horizontal 2-dimensional plane (possibly free of obstacles such as furniture and stairs), heavily limiting the movement space of the experiment.

\begin{figure}
	\begin{center}
	\includegraphics[width=0.48
	\textwidth]{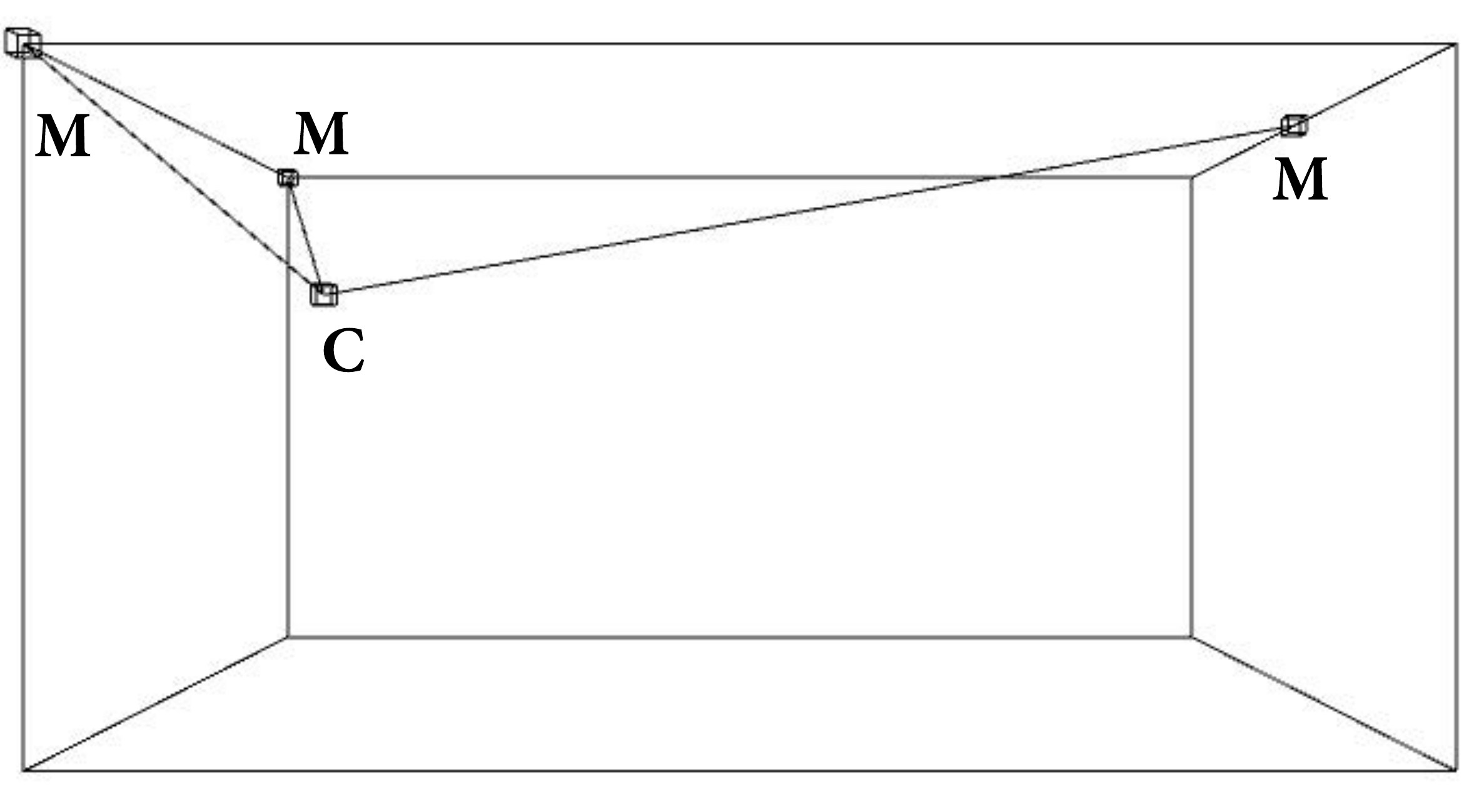}
	\caption{Example of Gondola infrastructure in a 3-dimensional space. In this example the carriage (C) is attache to three spooling motors on an office ceiling.}
	\label{fig:3d}
	\end{center}
\end{figure}

To avoid the aforementioned problems, we present the design of Gondola, a robotic infrastructure that moves through cables, rather than wheels. Inspired by plotters based on polar coordinates~\cite{DanRoyer,JurgLehni202,SandyNoble} our robotic system embeds the mobile wireless sensor in a carriage, which is connected trough thin wires to one or more spooling motor, depending on the required degree-of-freedom of the movement (see Fig.~\ref{fig:3d}).

Because the design of Gondola is completely parametric (both the location and the number of spooling motors), it can be easily adapted to different environments (small rooms, halls, outdoor) and needs (linear motion, volumetric scannings). Moreover, because its movement is not bounded to the ground plane, Gondola is less affected by obstacles than traditional wheeled robots. 

Preliminary experiments show that, in a 6.5$\times$3.9$\times$3.1 meter room, Gondola repeatedly achieves a positioning error of less than 2\,cm. This error can be further reduced with a proper design of the spooling mechanism, a topic we briefly discuss in Section~\ref{sec:discussion}.

Finally, Gondola parametric infrastructure is completely open-source. The design files of both hardware and software are available at \url{http://github.com/iprotonotarios/gondola}. 

\section{The Gondola Platform}
\label{sec:mechanism}

\begin{figure}
	\includegraphics[width=0.48
	\textwidth]{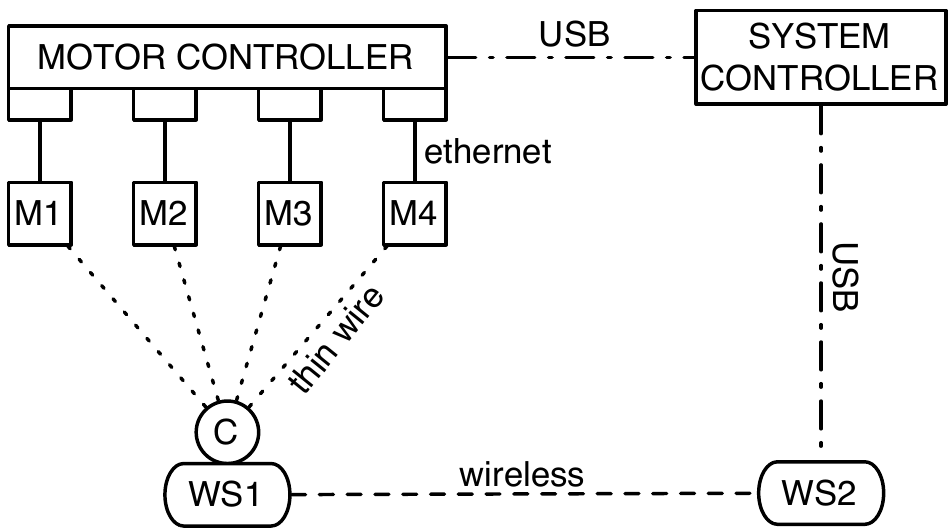}
	\caption{The system architecture of Gondola. The carriage (C) is connected to one or more spooling motors (M) via thin wires. By controlling the length of the wire spooled by each motor, the Motor and System controllers are able to move a carried object (WS) in space.}
	\label{fig:architecture}
\end{figure}

The architecture of Gondola, shown in Fig.~\ref{fig:architecture}, is composed of several modules. 
(i) The \emph{System Controller}, which gets a sequence of 3-dimensional positions (carriage's trace) and translate them into a sequence of 1-dimensional spooling movements. One for each spooling motor. 
(ii) The \emph{Motor Controller}, which is in charge of receiving a spooling sequence and properly actuate the motors such that the resulting movement in the 3-dimensional space is smooth and the speed is constant.
(iii) The \emph{Carriage} (C), which is connected to each motor via thin wires and carries devices such as a wireless sensors (WS1). Once the Carriage (C) reaches the intended position, the System Controller logs the experiment output running on another wireless device (WS2) until an event occurs. Then, the Carriage is moved to the next scripted location.

We now analyze in detail the implementation details of our system.

\subsection{System Controller}
Gondola's System Controller comprise of a computer running a software that interfaces with the user (who runs an experiment), a wireless sensor node (WS2) and the Motor Controller. Its interface is written in Processing\footnote{\url{https://processing.org}} and can run on several platforms. 

Through this interface a user can input the 3D coordinates of the location of each spooling motor, together with the carriage's starting position. This calibration step is essential to convert a desired movements in 3D space to 1D spolling distances. 
In particular, given a 3D movement from point A to point B, the 1D spolling distance of a motor M is computed as 
$$\text{dist}(A,M) - \text{dist}(B,M),$$ 
where $\text{dist()}$ computes the euclidean distance between two points in the 3D space. 
Note that, theoretically, Gondola works with any motor configuration \eg number of motors, position of motors. Nevertheless, because the pulling wires are tensioned only by gravity, to achieve a good range of movement, Gondola needs at least three properly positioned motors.  

Once the required spooling distance is computed for each motor, the System Controller sends a command to the Motor Controller and waits for an acknowledgment (ACK), indicating that gondola reached the required position. 
Then, the System Controller starts logging the experiment output until a predefined event occurs \eg a timeout or a specific output from the sensor node, and the System Controller can proceed processing the next planned position (if any).

\subsection{Motor Controller}
\begin{figure}
	\includegraphics[width=0.48
	\textwidth]{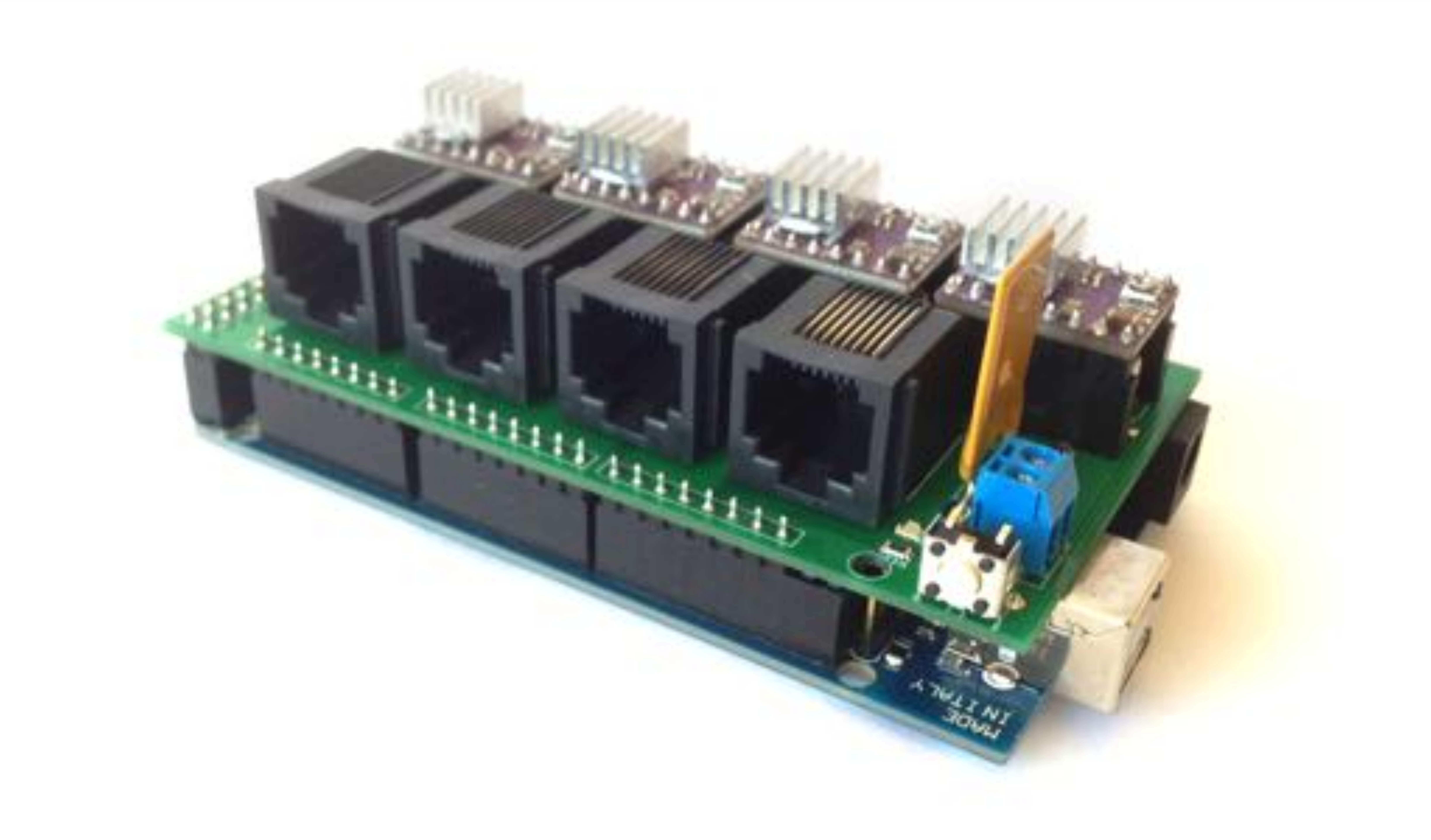}
	\caption{Gondola's Motor Controller (green pcb). The shield hosts an ethernet interface and a stepper drivers (violet pcb) for each of the four spooling motors. Computation and serial communication is performed by the underneath Arduino Mega (blue pcb).}
	\label{fig:motor}
\end{figure}

The Motor Controller is a combination of hardware and software that drives a series of motors, each one precisely spooling/unspooling a pulling wire according to the instructions received by the System Controller. For simplicity and cost effectiveness, we designed the Motor Controller as a shield for Arduino Mega\footnote{\url{http://www.arduino.cc/en/Main/ArduinoBoardMega2560}} that can interface with up to four stepper motors via ethernet cables. 

Note that the current version of Gondola relies only on the stepper motors to control the spooled length of the pulling wires, with no feedback control. Thus, it requires a constant calibration to keep an unbiased information of length actually spooled by each motor. 
To overcome this problem, it is possible in the future to add a rotary encoder to each one of the spooling motors and provide feedback to our system. The Motor Controller allows such modification, interfacing four of the eight ethernet wires with the Arduino's GPIO (the other four wires are used to control the stepper motor). 

\subsection{Spooling Motor(s)}
\begin{figure}
	\includegraphics[width=0.48
	\textwidth]{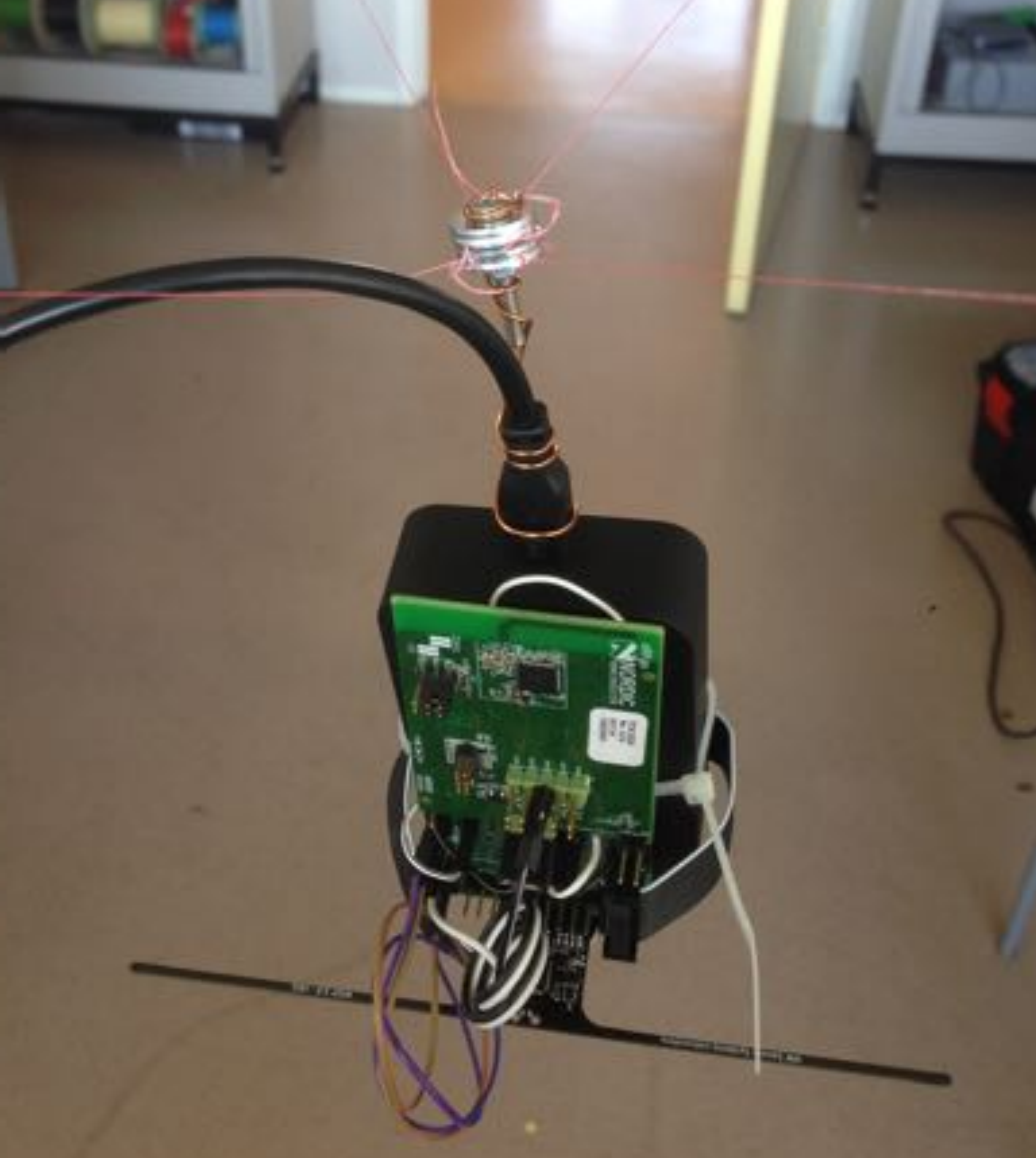}
	\caption{Gondola's moving an heavy power supply and a wireless device.}
	\label{fig:gondola}
\end{figure}

Our robotic infrastructure moves the Carriage (and the wireless sensor) by changing the length of wire, spooled by each motor. 
The characteristics of these motors are therefore very important for the overall performance of the system. 
While it is obvious that movement precision is important, it is not obvious that other characteristics, such as the holding torque (the capacity to maintain a position, when the carriage is loaded), are equally important (see Fig.~\ref{fig:gondola}). 
Moreover, to improve usability, motors must spool fast and smoothly.

For our implementation, we choose four 42BYGHW811 Wantai stepper motors, with 1.8$^{\circ}$ movement precision and a holding torque of 4800 g-cm. Because each motor drive a wheel of radius 2\,cm (no gears), the resulting movement precision is ($\pi$ 4)1.8/360 = 0.062\,cm, while the holding force is 2400\,g. 
Note that, because of the limited diameter chosen in our implementation, the pulling wire will spool several times around the spooling wheel, changing its radius and, thus, the spooled length, speed and force. As we will see in the evaluation section, this will affect the precision of our system.
To reduce the aforementioned problem, each spool used a special 0.01\,cm-thin fishing line with minimal elasticity and capable of holding up to 7000\,g. 

\section{Evaluation}
\label{sec:evaluation}

\begin{figure}[t]
\begin{center}
	\subfloat[Gondola's positioning error for linear 1-dimensional movements. The less wire motors spool, the higher the relative error.]{
	\includegraphics[width=0.48
	\textwidth]{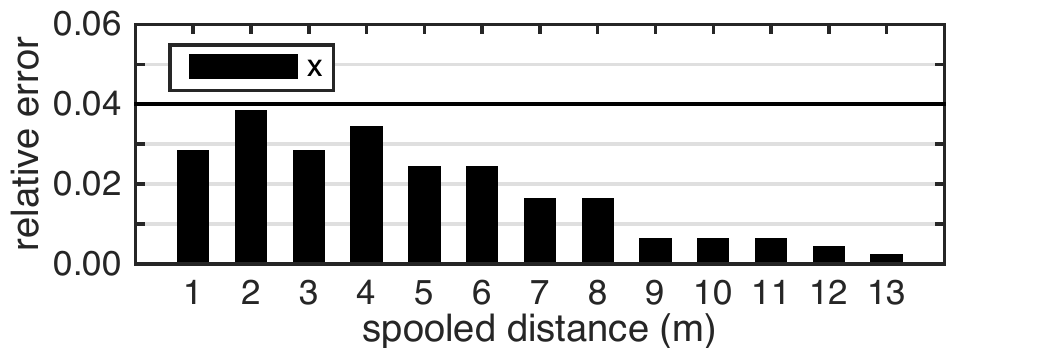}
	\label{fig:linear_error}
	}\\	
	\subfloat[Gondola's positioning error for spatial 3-dimensional movements. The further Gondola moves from the center of the room, the higher the relative error.]{
	\includegraphics[width=0.48
	\textwidth]{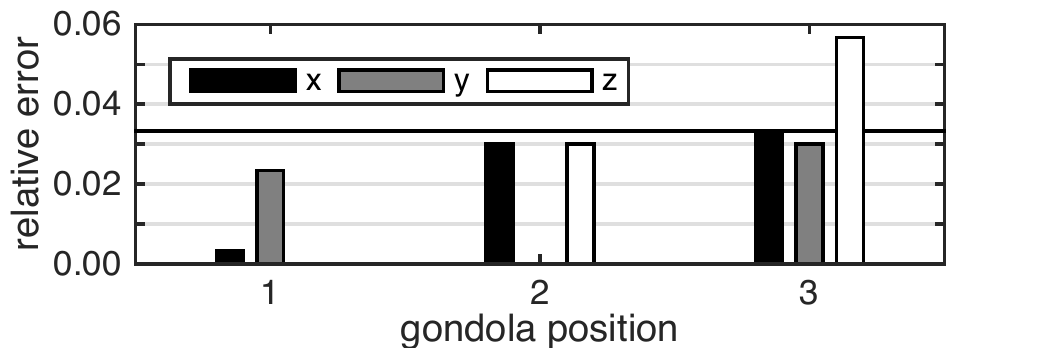}
	\label{fig:spatial_error}
	}
	\end{center}
\caption{Gondola relative positioning error in 1- and 3-dimensional spaces. For comparison, the horizontal black line indicates the relative error produced by an absolute movement error of 1\,cm.}
\label{fig:error}
\end{figure}

\begin{table}
	\begin{center}
\begin{tabular}{|c|ccc|ccc|}
\hline
\multirow{2}{*}{Position} & \multicolumn{3}{c|}{Absolute} & \multicolumn{3}{c|}{Normalized}\\ \cline{2-7}
 & x & y & z & x & y & z\\
\hline
\hline
\multicolumn{7}{|c|}{Linear Experiments}\\
\hline
1 & 50 & - & - & 0.07 & - & -\\
2 & 100 & - & - & 0.14 & - & -\\
3 & 150 & - & - & 0.21 & - & -\\
4 & 200 & - & - & 0.28 & - & -\\
5 & 250 & - & - & 0.35 & - & -\\
6 & 300 & - & - & 0.42 & - & -\\
7 & 350 & - & - & 0.49 & - & -\\
8 & 400 & - & - & 0.56 & - & -\\
9 & 450 & - & - & 0.63 & - & -\\
10 & 500 & - & - & 0.70 & - & -\\
11 & 550 & - & - & 0.77 & - & -\\
12 & 600 & - & - & 0.84 & - & -\\
13 & 650 & - & - & 0.91 & - & -\\
\hline
\hline
\multicolumn{7}{|c|}{Spatial Experiments}\\
\hline
1 & 355 & 196 & 310 & 0.54 & 0.50 & 0.99\\
2 & 405 & 86 & 240 & 0.61 & 0.22 & 0.77\\
3 & 495 & 196 & 240 & 0.75 & 0.50 & 0.77\\
\hline
\end{tabular}
\end{center}
\caption{Starting positions for the linear and spatial experiments.}
\label{tab:positions}
\end{table}

In order to evaluate the characteristics of Gondola, we measured the accuracy of each individual motor (1D linear movement) and later, the overall system (3D spatial movement). In particular, we set the position of Gondola to a set of starting coordinates (summarized in Table~\ref{tab:positions}) and measured the relative error for a fixed-length movement (25 and 30\,cm for the linear and spatial movements, respectively).

In the case of the linear spooling distance, the results in Figure~\ref{fig:linear_error} shows that the more wire is spooled, the smaller the error. This is due to the fact that, when lots of wire is spooled, the diameter of the spooling wheel increases, enlarging the spooled lengths.

This affects in order Gondola's movement in 3-dimensional space. As soon as Gondola is positioned in the center of the experimental room, all the motors' spooled distance are long (position 1, Fig.~\ref{fig:spatial_error}) and Gondola's position error in space is very low. 
As soon as Gondola moves towards one angle of the room (position 2 and 3, Fig.~\ref{fig:spatial_error}), few motors' spooled distance reduce drastically, increasing the linear positioning error and, thus, the spatial error in 3-dimensional space.

\section{Discussion}
\label{sec:discussion}
In this paper we presented Gondola, a parametric robotic system that provides an accurate and repeatable movements for Wireless Sensor Networks. Thanks to its flexibility, Gondola can be easily adapted to different environment and testing scenarios, from linear movements (using only 1 motor) to 3-dimensional movements (using 3 or more motors).

Nevertheless, accurately spool the desired wire length has proven to be one of the main challenges of Gondola. In the future, we plan to explore different solution to overcome this problem. From adding a feedback loop, based on rotary encoders, to substitute the actual simple wires (fishing lines) with ball-chain wires. We argue that precisely spooling the desired length of wire, together with an accurate measurement of the motors' position are the keys to improve even more the positioning accuracy of Gondola.


\bibliographystyle{abbrv} 
 {
 \small
 \bibliography{library}
 }


\end{document}